\documentclass[10pt,conference]{IEEEtran}
\IEEEoverridecommandlockouts
\usepackage{cite}
\usepackage{amsmath,amssymb,amsfonts}
\usepackage{algorithmic}
\usepackage{graphicx}
\usepackage{textcomp}
\usepackage{xcolor}
\usepackage{gensymb}
\usepackage{multirow}

\def\BibTeX{{\rm B\kern-.05em{\sc i\kern-.025em b}\kern-.08em
    T\kern-.1667em\lower.7ex\hbox{E}\kern-.125emX}}

\newcommand{\headingstyle}[1]{\textbf{#1.}}

\begin{document}

\title{Non-Uniform Interpolation in Integrated Gradients for Low-Latency Explainable-AI

\thanks{ Accepted to appear in 56\textsuperscript{th} IEEE International Conference on Circuits and Systems (ISCAS 2023)}

}



\author{\IEEEauthorblockN{Ashwin Bhat, Arijit Raychowdhury}
\IEEEauthorblockA{\textit{School of Electrical and Computer Engineering},
\textit{Georgia Institute of Technology},
Atlanta, GA, USA}
ashwinbhat@gatech.edu, arijit.raychowdhury@ece.gatech.edu}

\maketitle

\begin{abstract}
There has been a surge in Explainable-AI (XAI) methods that provide insights into the workings of Deep Neural Network (DNN) models. Integrated Gradients (IG) is a popular XAI algorithm that attributes relevance scores to input features commensurate with their contribution to the model's output. However, it requires multiple forward \& backward passes through the model. Thus, compared to a single forward-pass inference, there is a significant computational overhead to generate the explanation which hinders real-time XAI. This work addresses the aforementioned issue by accelerating IG with a hardware-aware algorithm optimization. We propose a novel non-uniform interpolation scheme to compute the IG attribution scores which replaces the baseline uniform interpolation. Our algorithm significantly reduces the total interpolation steps required without adversely impacting convergence. Experiments on the ImageNet dataset using a pre-trained InceptionV3 model demonstrate \textit{2.6-3.6}$\times$ performance speedup on GPU systems for iso-convergence. This includes the minimal \textit{0.2-3.2}\% latency overhead introduced by the pre-processing stage of computing the non-uniform interpolation step-sizes. 
\end{abstract}

\begin{IEEEkeywords}
Explainable AI (XAI), Deep Neural Networks (DNN), Hardware-Aware Algorithm Design, GPU systems
\end{IEEEkeywords}

\IEEEpeerreviewmaketitle

\section{Introduction}
There has been a massive growth in the field of Machine Learning (ML) and Artificial Intelligence (AI). However, the black-box nature of DNN models has hindered its ubiquitous utilization \cite{guidotti2018survey}. Explainable-AI (XAI) provides insights into the workings of these models to enable adoption in safety-critical tasks \cite{Adadi2018} which require transparency and interpretability \cite{roscher2020explainable}. 
Within the field of XAI, feature attribution methods generate an explanation by scoring input features proportional to their contribution to the network's output \cite{Ancona2017}. For image based applications, these relevance scores are visualized as a heatmap \cite{ras2022explainable}. These post-hoc techniques can be applied to existing pre-trained models \cite{Samek2021}. Integrated Gradients (IG), a feature attribution algorithm, has become popular thanks to its ease of implementation, axiomatic theoretical underpinnings, and applicability to any differentiable model \cite{sundararajan2017axiomatic}.

IG accumulates gradients along a straight-line interpolation path between a baseline and the input. A baseline is indicative of missingness or lack of input \cite{sturmfels2020visualizing}. For example, a black image is a commonly used baseline for vision tasks. IG requires multiple forward (inference) and backward (gradient backpropagation) passes through the model for each input. Thus, there is a large computational overhead in generating the explanation (50-1000$\times$ slower) compared to just evaluating the model's output which required a single forward pass. As noted in this XAI deployment study \cite{bhatt2020explainable}, it is necessary to reduce this overhead to enable real-time low-latency XAI. It is vital to overcome the technical limitations of computing explanations quickly in domains like smart healthcare \cite{ibrahim2020explainable}, medicine \cite{tjoa2020survey}, finance \cite{hadji2021explainable}, and hardware security \cite{golder2022exploration} where IG is being utilized.    

Several optimizations over baseline IG have been proposed to improve the quality of the generated heatmaps. \cite{sturmfels2020visualizing} proposes averaging the attributions obtained by using several different baselines. Google's XRAI \cite{Kapishnikov_2019_ICCV} segments the input into several regions and applies IG on each segment before stitching the results together. Captum, a PyTorch based XAI library developed by Meta \cite{kokhlikyan2020captum}, uses Noise Tunnel which averages the IG attributions over several noisy copies of the original input \cite{smilkov2017smoothgrad}. Despite using baseline IG multiple times in their pipeline, none of these algorithms attempt to reduce its computational overhead. Thus, they stand to gain significant performance benefits from an IG implementation optimized for low-latency on the underlying hardware platform.

\begin{figure}[ht!]
    \centering
    \includegraphics[width=1\linewidth]{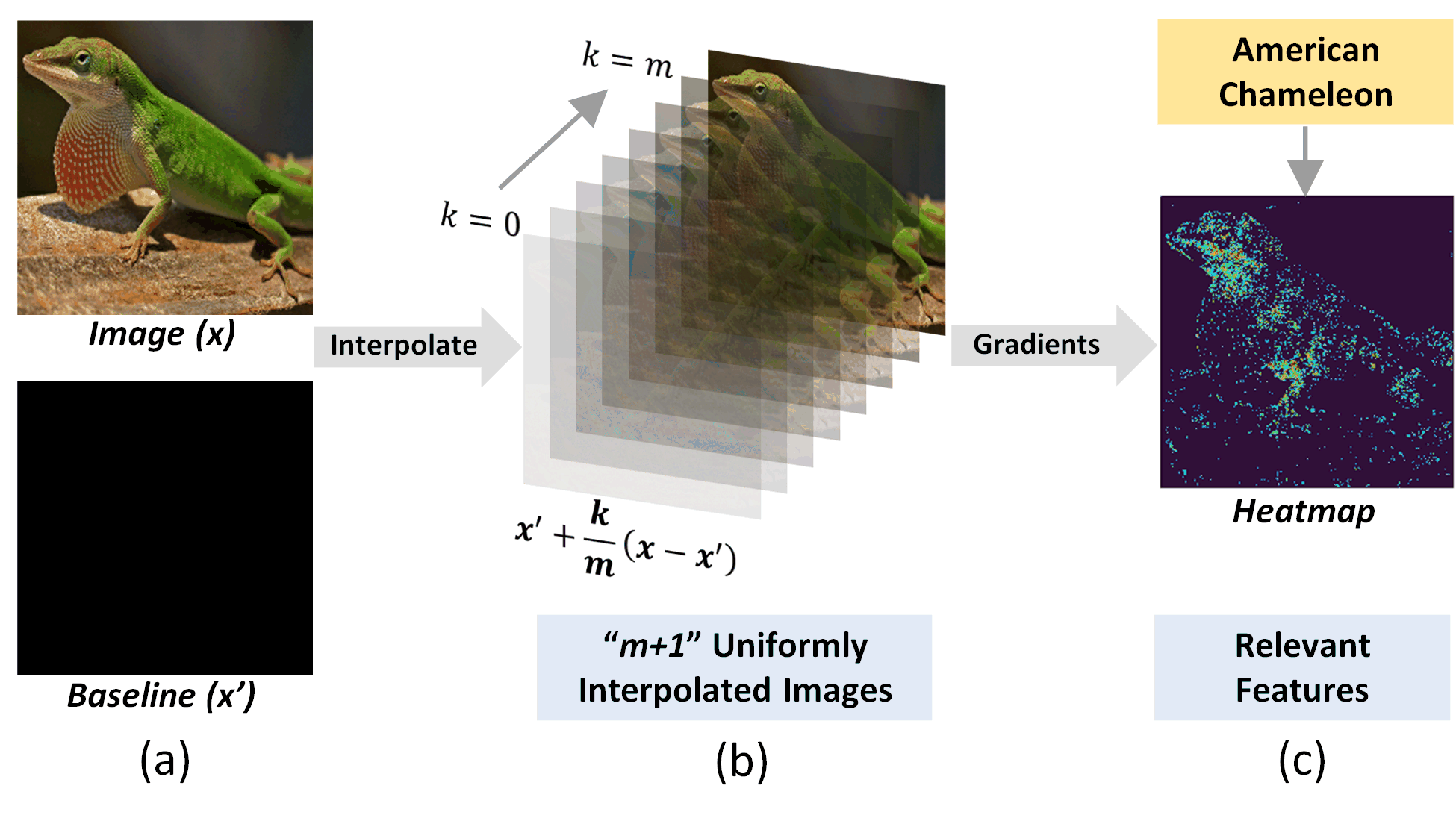}
    \caption{Overview of Feature Attribution using the Integrated Gradients (IG) algorithm (a) Inputs to the algorithm: Image ($x$) and Baseline ($x'$) (b) Interpolated images along straight line path between Baseline ($k=0$) and Image ($k=m$) (c) Visualization of accumulated gradients to highlight input features relevant for classifying image as American Chameleon (target class)}
    \label{fig:IG overview}
\end{figure}

All the currently known implementations of IG use uniform interpolation between baseline and input to approximate the continuous IG integral [Eq. \ref{eq:IG Integral}] as Riemann-sum [Eq. \ref{eq:IG Discrete}]. In this work, we propose a non-uniform interpolation scheme to reduce the discretization steps required while maintaining similar convergence accuracy. We first identify regions in the IG path with higher information content. By having smaller step-size in these regions and larger step-size outside of them, the overall compute overhead of IG is reduced.

In summary, this work makes the following contributions:
\begin{itemize}
    \item To the best of our knowledge, our work is the first of its kind to design and employ a non-uniform interpolation scheme along the IG path to compute the discrete Riemann-sum approximation of the IG integral. 
    \item Based on the experimental observations, we propose and justify the use of change in classification probability along the IG path as an information content metric to determine the non-uniform discretization step-size.
    \item We demonstrate the performance improvement achieved on GPU systems and quantify the latency overhead introduced by our algorithm compared to the baseline IG which uses a uniform interpolation scheme.
\end{itemize}

\section{Integrated Gradients (IG)}
The Integrated Gradients (IG) algorithm is a feature attribution method [Fig. \ref{fig:IG overview}]. Formally, for a given input ($x$) and a model function ($f$), a feature attribution method assigns a relevance score ($\phi _i(x,f)$) to the i\textsuperscript{th} input feature. The score is a measure of that feature's contribution to the model's output.

The simplest way to assign a relevance score is to evaluate the gradient of the model's output with respect to the input feature. A large gradient value implies that small changes in the feature value produce a large change in the model's output, thereby, indicating higher relevance. However, gradients are a local explanation method that can suffer from saturation effects. Path attribution methods (PAM) overcome this issue by accumulating gradients along a path between a baseline and the actual input \cite{lundstrom2022rigorous}. These are mathematically motivated and satisfy desirable properties such as completeness and sensitivity. IG is a subset of PAM which considers a straight-line path between the baseline ($x'$) and the input ($x$) as shown in Eq. \ref{eq:IG Integral}. The commonly used baselines for computer vision applications include black, white, or random noise images \cite{sturmfels2020visualizing}. They represent the notion of missingness or lack of any input.  

\begin{equation}
    \label{eq:IG Integral}
    \phi _i\left(f,x,x'\right) = (x - x') \times \int_{0}^{1} \frac{\partial f(x'+\alpha(x-x'))}{\partial x_i} \,d\alpha 
\end{equation}

In Eq. \ref{eq:IG Integral}, $f$ is the model function and $\alpha$ is the interpolation constant along the straight-line path. In practice, to evaluate IG attributions, the continuous integral is approximated as a Riemann sum with \textit{$m+1$ uniform steps} as shown in Eq.\ref{eq:IG Discrete}.

\begin{equation}
    \label{eq:IG Discrete}
    \phi _i\left(f,x,x'\right) = (x - x') \times \frac{1}{m} \sum_{k=0}^{m} \frac{\partial f(x'+\frac{k}{m}(x-x'))}{\partial x_i}
\end{equation}


The number of steps ($m$) typically ranges from 200 to 1000 \cite{sotoudeh2019computing}. Its value is chosen based on the convergence metric ($\delta$) which is defined (Eq. \ref{eq:Convergence}) using the completeness property \cite{lundstrom2022rigorous} satisfied by the continuous integral formulation of IG. 

\begin{equation}
    \label{eq:Convergence}
    \delta = \left| \sum_{i} \phi_i(f,x,x') - [f(x) - f(x')] \right|
\end{equation}

From Eq. \ref{eq:IG Discrete}, we observe that for every input, the IG algorithm creates several interpolated versions [Fig. \ref{fig:IG overview}(b)]. It then computes the gradient of the model's output with respect to input features for each one. This step requires a forward and a backward pass through the model. These gradients are then aggregated to assign the overall attribution score [Fig. \ref{fig:IG overview}(c))].

\section{Methodology}

\begin{figure}[ht!]
    \centering
    \includegraphics[width=1\linewidth]{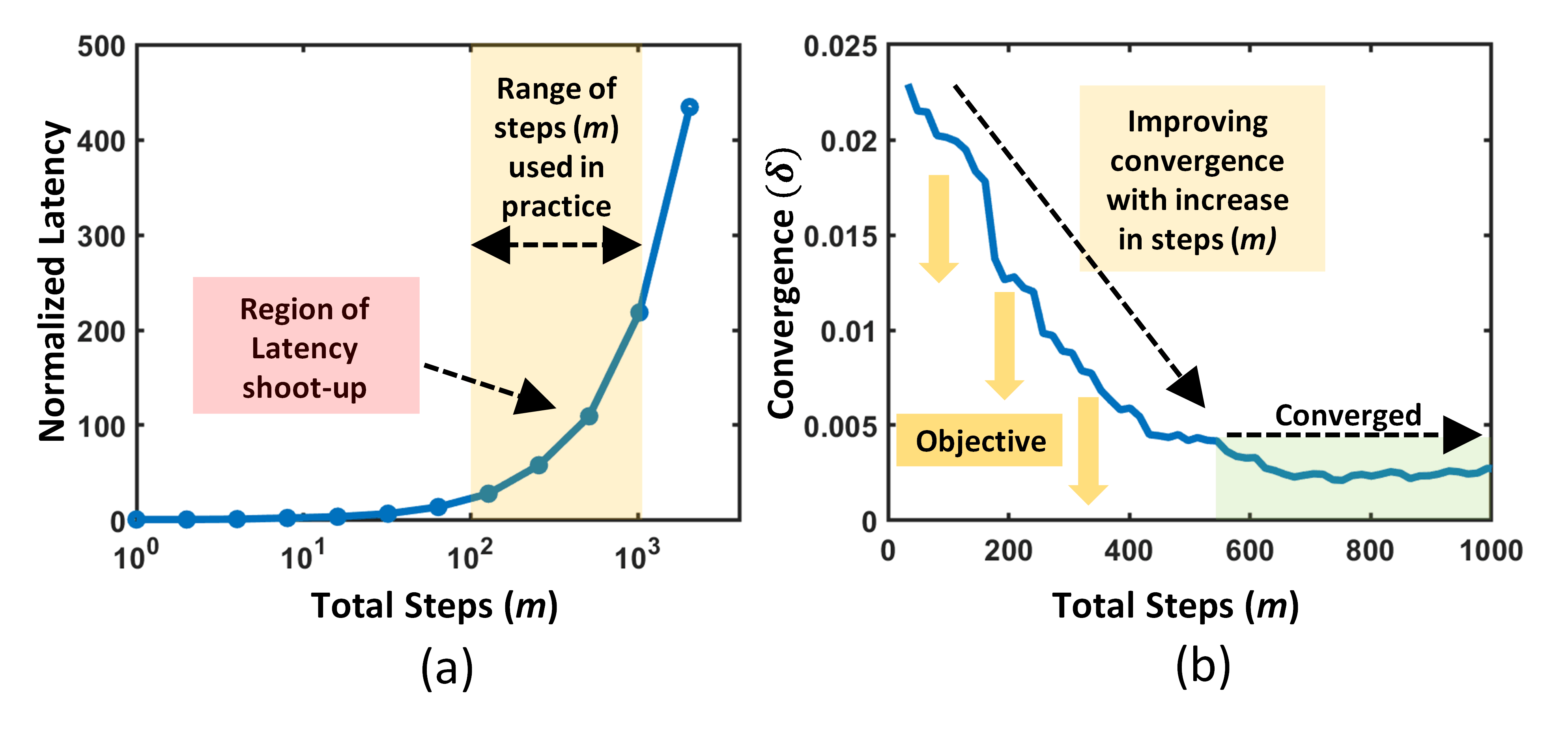}
    \caption{(a) Latency increases with increase in number of interpolation steps [values normalized relative to the latency for m=1] (b) Decreasing convergence $\delta$ with increasing steps implies better convergence at reduced interpolation step-size. Thus, there is a latency penalty for better convergence.}
    \label{fig:background}
\end{figure}

\headingstyle{Background} 
The run-time latency of IG depends on the number of interpolation steps [Fig. \ref{fig:background}(a)]. More steps yield better convergence $\delta$ [Fig. \ref{fig:background}(b)] at the cost of higher latency.

\headingstyle{Objective} 
The goal of this work is to reduce the compute overhead of IG for generating explanations. The typically used range of values for the number of steps lies beyond the knee-point of the latency v/s step-count graph [Fig. \ref{fig:background}(a)]. Thus, for lower latency, the number of interpolation steps must be decreased without compromising convergence [Fig. \ref{fig:background}(b)].

\begin{figure}[ht!]
    \centering
    \includegraphics[width=1\linewidth]{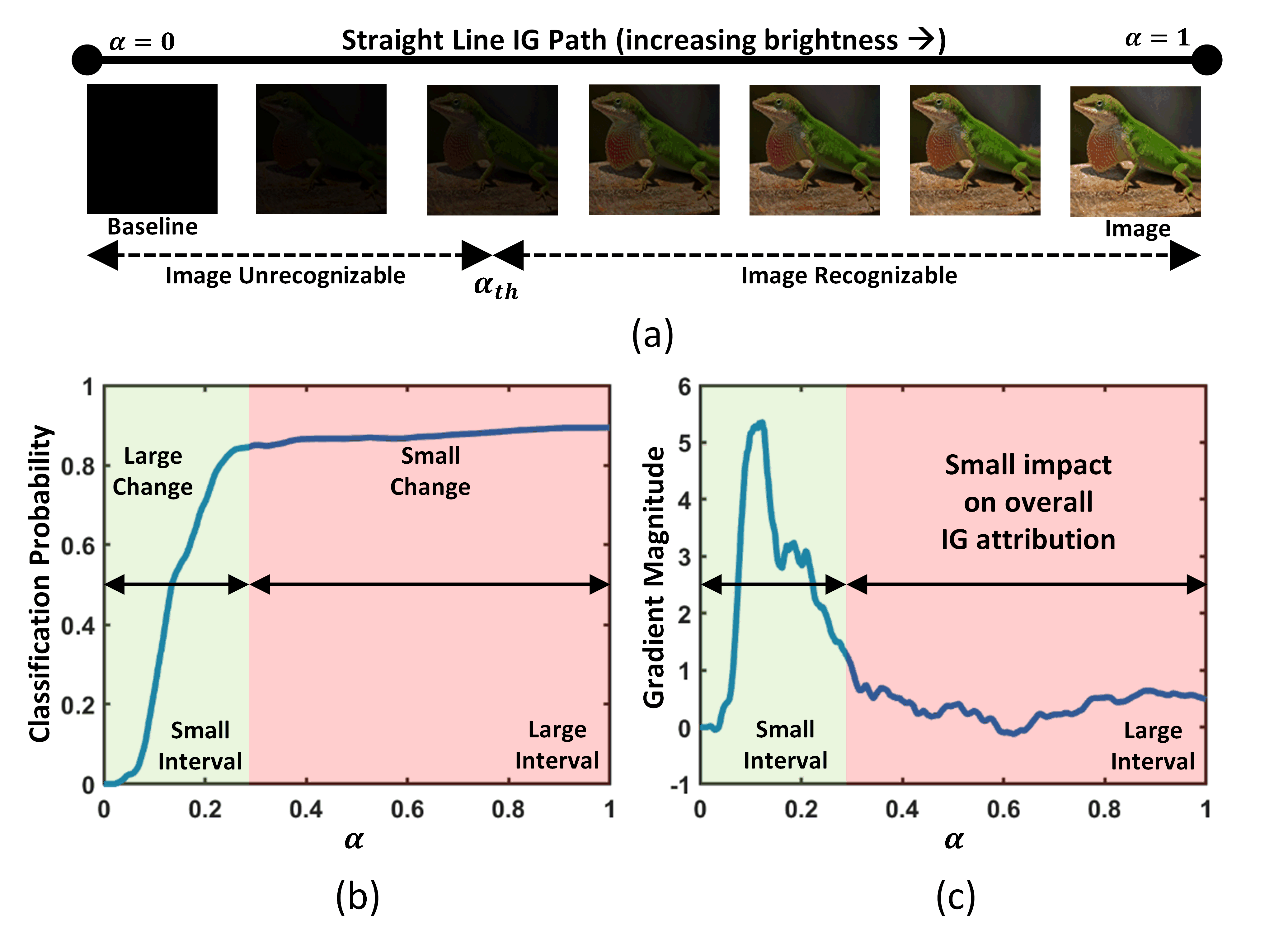}
    \caption{Variation along the IG path of (a) Distinguishability of the image from its interpolated copies (b) Classification probability of the model (c) Contribution to the convergence term based on the relative gradient magnitude. Thus, a small region along the IG path contains most of the information.}
    \label{fig:observation}
\end{figure}

\headingstyle{Observation}
Along the interpolation path, close to the baseline (small values of $\alpha$), the actual input image is unrecognizable from its interpolated version. However, after a certain $\alpha$, the input can be identified. Beyond this threshold, as we move towards the input along the IG path, the brightness of the interpolated image increases. Intuitively, for a human observer, the change in classification confidence is not uniform along the IG path as seen in Fig. \ref{fig:observation}(a). 

We test this intuition on the model. The classification probability of the model changes sharply as we increase the interpolation constant $\alpha$ [Fig. \ref{fig:observation}(b)] along the IG path. At $\alpha=0.25$, the classification probability (0.83) is $>$90\% of its final value (0.89) for the input image ($\alpha=1$). Thus, the model's confidence about the prediction is built over a small interval along the IG path with minimal change outside of it.

The IG algorithm accumulates the gradient of the classification probability with respect to input features for each interpolated image. Thus, the change in classification probability can be used as a metric for information content along the IG path. In regions of large change, the gradient values are also larger [Fig. \ref{fig:observation}(c)] and contribute more to the overall IG attribution. 

\begin{figure}[ht!]
    \centering
    \includegraphics[width=1\linewidth]{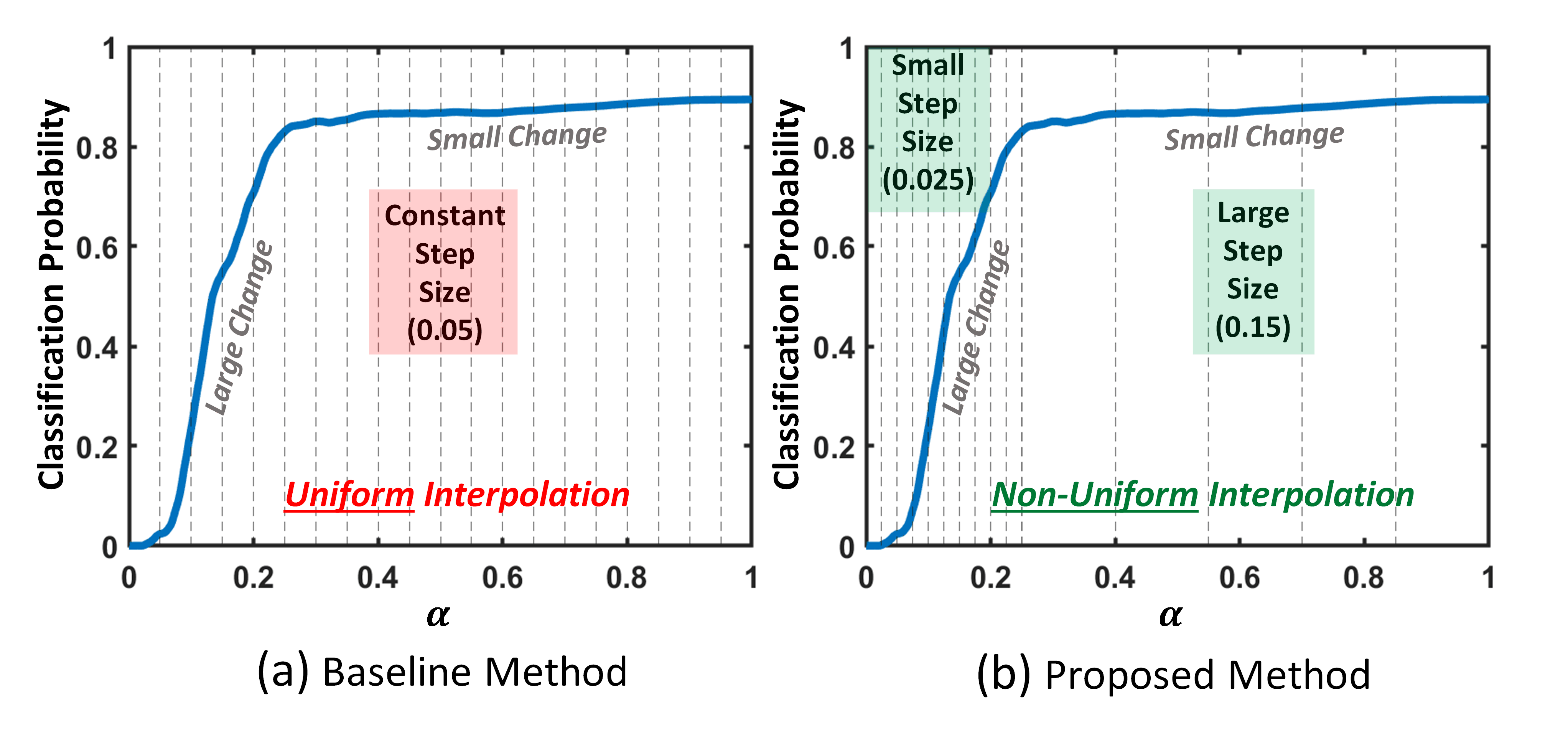}
    \caption{(a) Uniform interpolation with constant step size for the entire IG path (b) Non-Uniform Interpolation with small uniform step size in the region of large change and large uniform step size in the region of small change.}
    \label{fig:proposal}
\end{figure}

\headingstyle{Proposed Method}
We propose a non-uniform interpolation scheme to replace the baseline uniform interpolation [Fig. \ref{fig:proposal}(a)] along the IG path. Specifically, the IG path is divided into multiple intervals and uniform interpolation is performed within each interval using a different step-size. Based on the earlier observation [Fig. \ref{fig:observation}], we can use a smaller step size in the regions where there is a large change in the classification probability. Outside of such regions, a larger discretization step size can be used. Overall, the total steps are non-uniformly distributed along the IG path with a bias towards regions with higher information content.

\headingstyle{Algorithm}
The proposed algorithm comprises two stages. The \textit{first stage} takes the number of intervals ($n_{int}$) as a parametric input to divide the IG path into $n_{int}$ equal pieces. The classification probability of the interpolated images at the interval boundaries is evaluated to calculate its normalized change in each interval [$\Delta f(x_{int})$]. The total number of steps ($m$) is then distributed across each interval proportional to the square root of the change [$\sqrt{|\Delta f(x_{int})|}$]. We observed that linear dependence ($m_{int} \propto \Delta$) allotted negligible discretization steps to regions with small change. Hence, we use $m_{int}\propto\sqrt{\Delta}$ to attenuate the bias towards intervals of large change. In the \textit{second stage}, we perform uniform IG within each interval with the respective step count. The IG attributions of different intervals are then summed up to determine the overall IG attribution. The proposed algorithm is, therefore, \textit{uniform-in-intervals} but \textit{non-uniform overall} along IG path. Fig. \ref{fig:proposal}(b) illustrates this for $n_{int} =4$.   
\section{Results}

\headingstyle{Experimental Setup}
The proposed method can directly replace the baseline IG algorithm and be applied to any differentiable model. To demonstrate its efficacy, we test it on the ImageNet dataset \cite{deng2009imagenet} using a pre-trained InceptionV3 model \cite{szegedy2015going}. We consider our \textit{baseline} to be the existing IG implementation that employs uniform interpolation. We compare it against our proposed \textit{non-uniform interpolation algorithm} and vary the number of intervals ($n_{int}$) parameter.

\begin{figure}[ht!]
    \centering
    \includegraphics[width=1\linewidth]{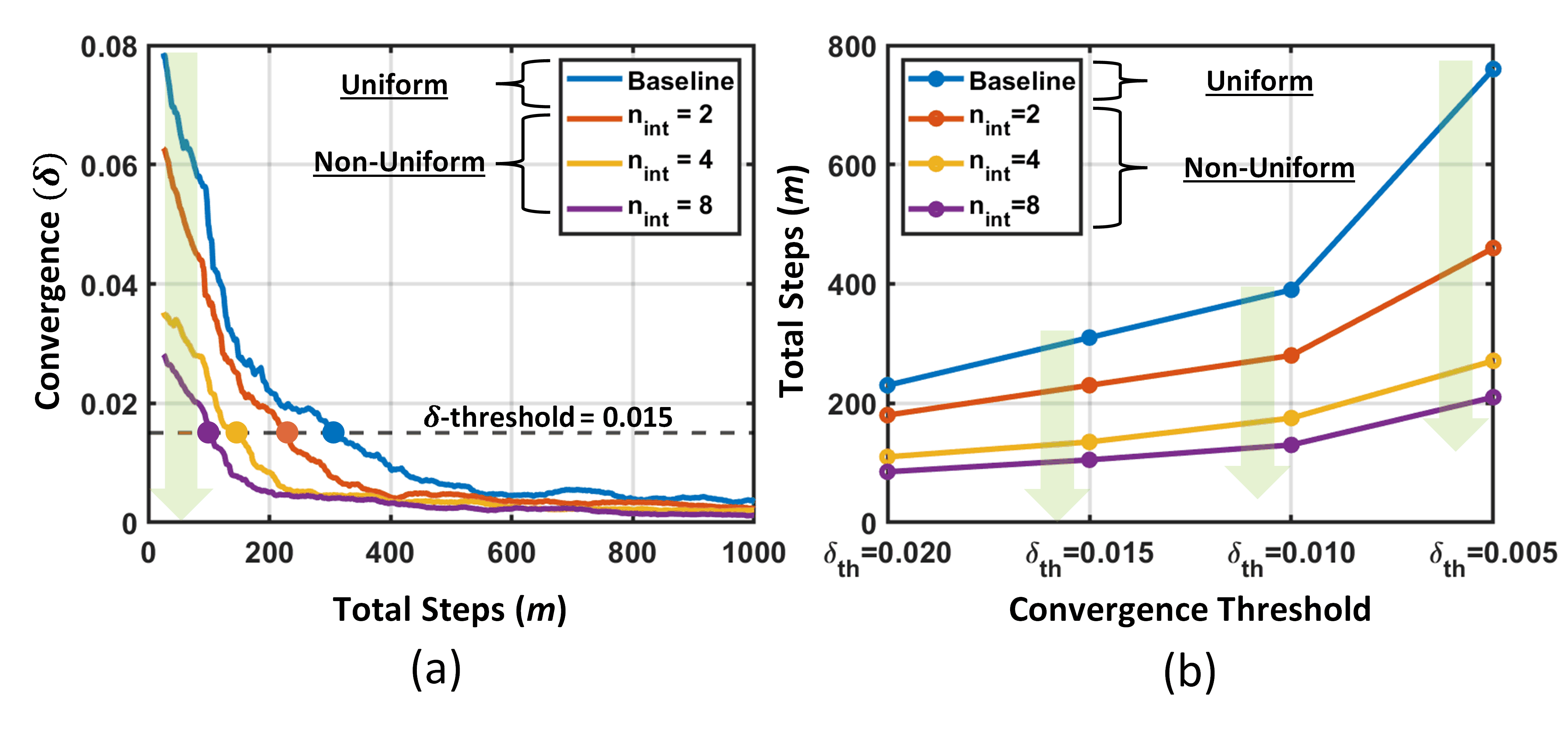}
    \caption{For different IG interpolation schemes, the variation of (a) Convergence delta ($\delta$) as we increase the total interpolation steps ($m$) (b) Total interpolation steps required for reaching the convergence threshold ($\delta_{th}$).}  
    \label{fig:result_conv_steps}
\end{figure}

\headingstyle{Convergence}
In Fig. \ref{fig:result_conv_steps}(a), we demonstrate the effect of the total number of steps on the convergence $\delta$ for different interpolation schemes. For any given number of steps ($m$), our proposed algorithm achieves better convergence $\delta$ compared to baseline. Thus, for iso-convergence, the proposed algorithm is able to reduce the total steps needed.

In practice, the total interpolation steps for IG are determined by fixing a threshold tolerance for the convergence. For example, in Fig. \ref{fig:result_conv_steps}(a), this threshold ($\delta_{th}$) is set to 0.015. The total number of steps is then chosen such that the convergence $\delta$ lies below the threshold [Fig. \ref{fig:result_conv_steps}(a)]. We vary $\delta_{th}$ and determine the number of steps required for different interpolation schemes [Fig. \ref{fig:result_conv_steps}(b)] to meet the convergence criterion. For all $\delta_{th}$ values, our proposed algorithm requires fewer steps and outperforms the baseline. Increasing the number of intervals further reduces the steps required for convergence. The benefits are more pronounced at smaller $\delta_{th}$ values. For $\delta_{th}=0.02$, we observe a 2.7$\times$ reduction while for $\delta_{th}=0.005$, we observe a 3.6$\times$ reduction in the total steps required for convergence.

We further observe that increasing the number of intervals ($n_{int}$) up to a certain point reduces $\delta$. Further increasing $n_{int}$ causes $\delta$ to increase since certain intervals are allotted negligible discretization steps which negatively impact convergence. Consequently, it increases the number of steps required to meet $\delta_{th}$. We observe that $n_{int} > 8$ manifests this issue. 

\begin{figure}[ht!]
    \centering
    \includegraphics[width=1\linewidth]{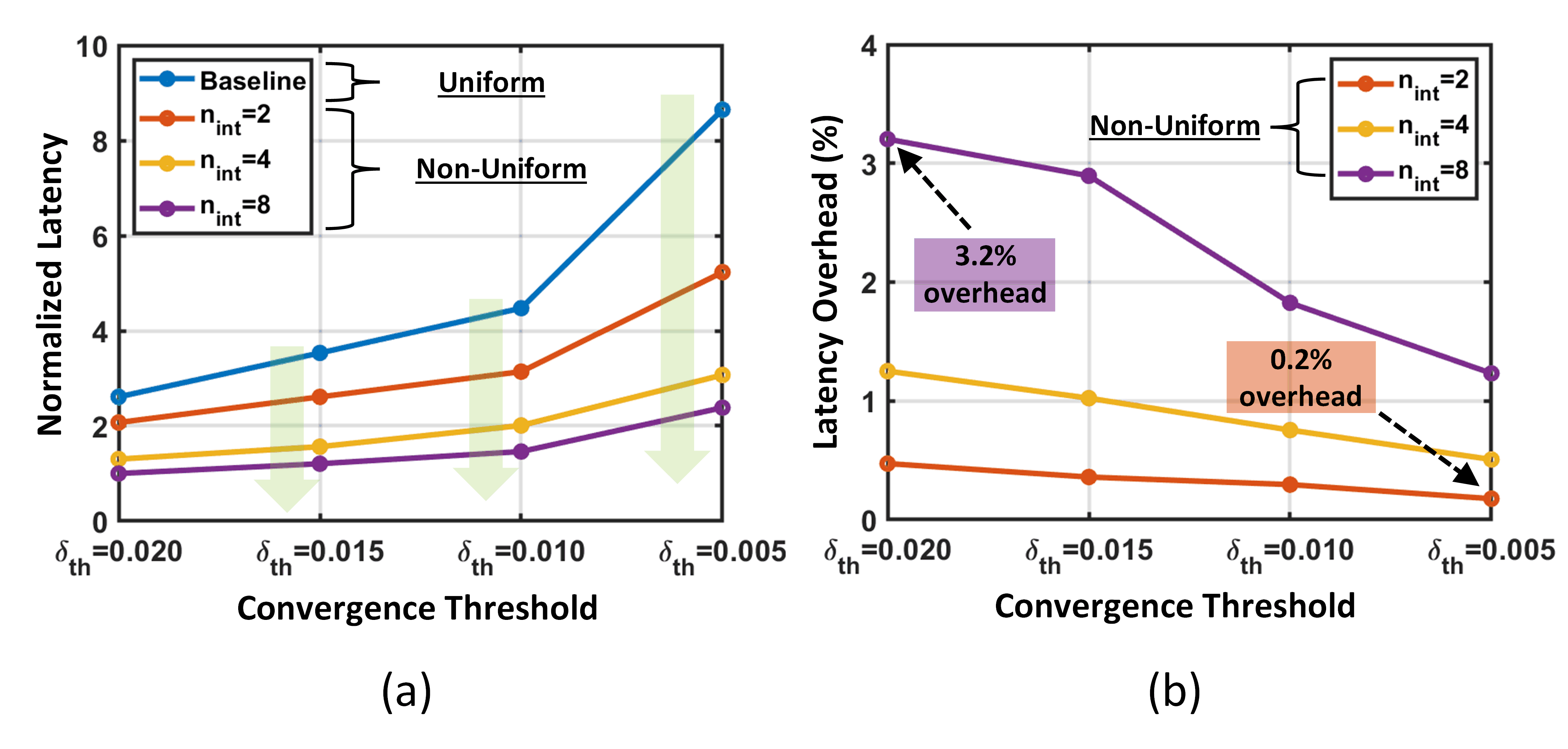}
    \caption{For each IG interpolation scheme, the variation of (a) normalized latency to meet the convergence threshold ($\delta_{th}$) (b) latency overhead (\% of total latency) of the first stage of the non-uniform interpolation algorithm. }
    \label{fig:result_latency}
\end{figure}

\headingstyle{Latency}
Since our method is applicable to any differentiable model and the performance benefit is not specific to a hardware architecture, we report normalized latency values. The latency is normalized relative to the algorithm configuration which yields the smallest run-time latency. In our experiments, we measure the run-time latency for running IG on InceptionV3 model with batch-size of 16 on a NVIDIA TITAN Xp GPU  using the PyTorch benchmark profiler. The profiler supports CUDA (excludes the overhead of thread synchronization), performs an initial warm-up, and averages over multiple runs to determine accurate execution latency.

The normalized latency [Fig. \ref{fig:result_latency}(a)] depends on the number of interpolation steps ($m$) which in turn depends on the convergence threshold ($\delta_{th}$). We make two observations. \textit{First}, the latency increases as we decrease $\delta_{th}$ because of an increase in $m$. However, the relative increase in latency as we reduce $\delta_{th}$ from 0.02 to 0.005 is higher for baseline (3.3$\times$) when compared to non-uniform interpolation (2.3-2.5$\times$). \textit{Second}, in terms of performance, our proposed non-uniform interpolation scheme outperforms the baseline across all $\delta_{th}$ values. The latency reduces as we increase $n_{int}$ yielding higher performance benefits. For $\delta_{th}=0.02$, we achieve a 2.6$\times$ and for $\delta_{th}=0.005$, we achieve a 3.6$\times$ latency reduction when compared to the baseline.

\headingstyle{Overhead} 
Determining the number of steps for each interval is the overhead of our proposed scheme. The memory overhead is minimal since we only store the classification probability of $n_{int}+1$ interpolated images to determine the step-sizes. The latency overhead is measured as a fraction of the total latency of running the IG algorithm [Fig. \ref{fig:result_latency}(b)]. The overhead varies between 0.2-3.2\% of the total latency.

We observe that the absolute value of the latency overhead depends only on the parameter $n_{int}$. This is because we run the inference pass through the network $n_{int}+1$ times to determine the classification probability change in each interval and distribute the total steps commensurately. Thus, the absolute overhead increases as we increase $n_{int}$. However, the relative value of the overhead depends on both $n_{int}$ and $\delta_{th}$  [Fig. \ref{fig:result_latency}(b)] because these parameters affect the total steps required and hence the overall latency of non-uniform IG.

\section{Discussion \& Related Work}
With XAI research being in its infancy, there is a paucity of work that focuses on improving performance (low-latency, high throughput, real-time) either through (a) specialized hardware architectures or (b) algorithmic optimizations with a hardware-aware design approach (this work). 

Pan et al.\cite{pan2022hardware} accelerates model-distillation based explanation on TPU or systolic-array hardware substrate. This method is unsuitable for low-latency explanation generation since it requires training a new model which locally mimics the input-output behavior of the black-box model for each input. In this work, we focus on a post-hoc explanation method that is directly applicable to any off-the-shelf differentiable model. Bhat et al.\cite{bhat2022gradient} accelerates gradient based heatmap visualization on FPGA platform. However, the implemented feature attribution algorithms suffer from local saturation effects. In this work, we focus on IG which overcomes this issue via path-attribution. Although we demonstrate our results on GPU systems, our algorithm is agnostic to the underlying hardware.

Sotoudeh et al.\cite{sotoudeh2019computing} proposes exact computation of the IG integral but achieves smaller performance gains (upto 1.7$\times$) compared to this work (upto 3.6$\times$). Kapishnikov et al.\cite{kapishnikov2021guided} avoids high-loss regions by updating a subset of features with low gradient magnitude at each point along the path. However, the next step is dynamically determined which limits the performance on GPUs as batch-size is restricted to 1. In our work, we design a static processing stage to pre-determine the discretization step size and leverage batching on GPU systems. Rahman et al.\cite{rahmanICLR} modifies the path by performing a local gradient ascent around each uniform interpolation point in the IG path. This magnifies the compute overhead of generating the explanation. Although both methods modify the baseline IG path to improve the quality of attribution heatmaps, they require more steps thereby incurring a performance overhead. Our work modifies the IG path to improve performance while maintaining iso-convergence with baseline IG. 

\section{Conclusion}
In this paper, a novel Non-Uniform Interpolation scheme for computing the Integrated Gradients attribution is presented. Our methodology is motivated by the observations we made on the baseline uniform interpolation. Regions of high-information content along the straight-line IG path are identified using the change in classification probability. Utilizing classification probability as an information metric is justified based on its correlation with gradient magnitude and its contribution to the overall IG attribution. The proposed algorithm distributes the steps among intervals with a bias towards ones with higher information content. Compared to the baseline, our algorithm meets iso-convergence thresholds with fewer total steps. We quantify the performance benefit on GPU systems using pre-trained models on the ImageNet dataset. Our experiments show that we can achieve a speed-up of \textbf{2.6-3.6}$\mathbf{\times}$ at a very low latency overhead of \textbf{0.2-3.2\%}. In summary, our hardware-aware algorithm design enables low-latency real-time explainable AI. 

\section{Acknowledgement}
This work was supported by Semiconductor Research Corporation (SRC) AI Hardware (AIHW) Task 2969.001 titled EXPERT: EXPlainable-AI through Efficient hardware design in EmeRging Technologies.

\bibliographystyle{IEEEtran}
\bibliography{refs}

\end{document}